\documentclass[conference]{IEEEtran}
\IEEEoverridecommandlockouts

\usepackage{cite}
\usepackage{mathtools}
\usepackage{amsmath,amssymb,amsfonts}
\usepackage{graphicx}
\usepackage{textcomp}
\usepackage{xcolor}
\usepackage{amsmath}
\usepackage{multirow}
\usepackage{multicol}
\usepackage{caption}
\usepackage{array}
\usepackage{subcaption}
\usepackage{balance}
\usepackage{wrapfig}
\usepackage{hyperref}
\usepackage{multirow}
\usepackage{multicol}
\usepackage{color}
\usepackage{parskip}
\usepackage{colortbl}
\usepackage{booktabs}
\usepackage{flushend}
\usepackage{hyperref}
\usepackage{todonotes}
\usepackage{comment}

\usepackage{cite}

\usepackage[utf8]{inputenc}
\usepackage[english]{babel}
\usepackage{enumitem}
\usepackage{algpseudocode}
\usepackage{algorithm}

\begin{document}

\title{Pain Forecasting using Self-supervised Learning and Patient Phenotyping: An attempt to prevent Opioid Addiction}


\author{\IEEEauthorblockN{Swati Padhee \IEEEauthorrefmark{1}, Tanvi Banerjee\IEEEauthorrefmark{1}, Daniel M. Abrams\IEEEauthorrefmark{2}}, and Nirmish Shah\IEEEauthorrefmark{3},
\IEEEauthorblockA{\IEEEauthorrefmark{1}Computer Science and Engineering, Wright State University, Dayton, OH, USA}
\IEEEauthorblockA{\IEEEauthorrefmark{2}Engineering Science and Applied Mathematics, Northwestern University, Evanston, IL, USA}
\IEEEauthorblockA{\IEEEauthorrefmark{3}Department of Medicine, Duke University, Durham, NC, USA}
}

\maketitle

\begin{abstract}
Sickle Cell Disease (SCD) is a chronic genetic disorder characterized by recurrent acute painful episodes. Opioids are often used to manage these painful episodes; the extent of their use in managing pain in this disorder is an issue of debate. The risk of addiction and side effects of these opioid treatments can often lead to more pain episodes in the future. Hence, it is crucial to forecast future patient pain trajectories to help patients manage their SCD to improve their quality of life without compromising their treatment. It is challenging to obtain many pain records to design forecasting models since it is mainly recorded by patients' self-report. Therefore, it is expensive and painful (due to the need for patient compliance) to solve pain forecasting problems in a purely supervised manner. In light of this challenge, we propose to solve the pain forecasting problem using self-supervised learning methods. Also, clustering such time-series data is crucial for patient phenotyping, anticipating patients' prognoses by identifying "similar" patients, and designing treatment guidelines tailored to homogeneous patient subgroups. Hence, we propose a self-supervised learning approach for clustering time-series data, where each cluster comprises patients who share similar future pain profiles. Experiments on five years of real-world datasets show that our models achieve superior performance over state-of-the-art benchmarks and identify meaningful clusters that can be translated into actionable information for clinical decision-making.
\end{abstract}


\section{Introduction}
\label{Intro}

Approximately half of the 117 million US adults are chronically ill and one in four has multiple chronic conditions (Ward et al. 2014). Patients with chronic diseases have to follow complex treatment regimens, making their care much more costly than that of a healthy individual. Not surprisingly, 86\% of all US healthcare spending is used to treat chronically ill patients \cite{gerteis2014multiple}. In this paper, we study Electronic Health Records (EHR) data from a cohort of patients with Sickle Cell Disease (SCD). SCD is a chronic lifelong illness and a disease multiplier, as it often goes hand-in-hand with other chronic conditions. Although severe acute pain episodes that correspond with vaso-occlusive events (VOEs) are the hallmark of SCD, a majority of patients with SCD report experiencing pain on most days \cite{knisely2023evaluating}.

Opioids are often used in the management of these painful episodes; the extent of its use in the management of pain in this disorder is an issue of debate. Some physicians advocate minimal use of opioid drugs for fear of addiction \cite{shapiro1997sickle,labbe2005physicians} while others believe that the under use of these medications in the control of pain may result in pseudoaddiction \cite{elander2004understanding,lusher2006analgesic}. There have also been reports in the literature of substance abuse by SCD patients \cite{boulmay2009cocaine,dohrenwend2005lost,alao2003drug}, though it is believed that the rate of drug abuse by SCD patients is not different from that of the general public \cite{meghani2012advancing}. In this paper, we address this issue by forecasting future pain trajectories for patients so that clinicians can plan a suitable and balanced pain management strategy, preventing risk for addictions.

 Current clinical standard involves patient self report for an estimate of pain which can be highly subjective. There can be a significant inter-individual variation in pain scores reported in response to the same stimulus. It can also be influenced by a confluence of factors such as mood and energy level at the time of pain report. Recent studies have demonstrated that machine learning models can be used to estimate these subjective pain levels from objective physiological signals \cite{panaggio2021can,padhee2020pain,yang2018improving,gruss2015pain,chu2017physiological,lopez2017multi}, facial expressions \cite{casti2019calibration,yang2018incorporating}, activity and motion tracking \cite{zamzmi2016approach,zamzmi2017review} have yielded promising results. 

\begin{figure*}
    \includegraphics[width=\textwidth,scale=0.7]{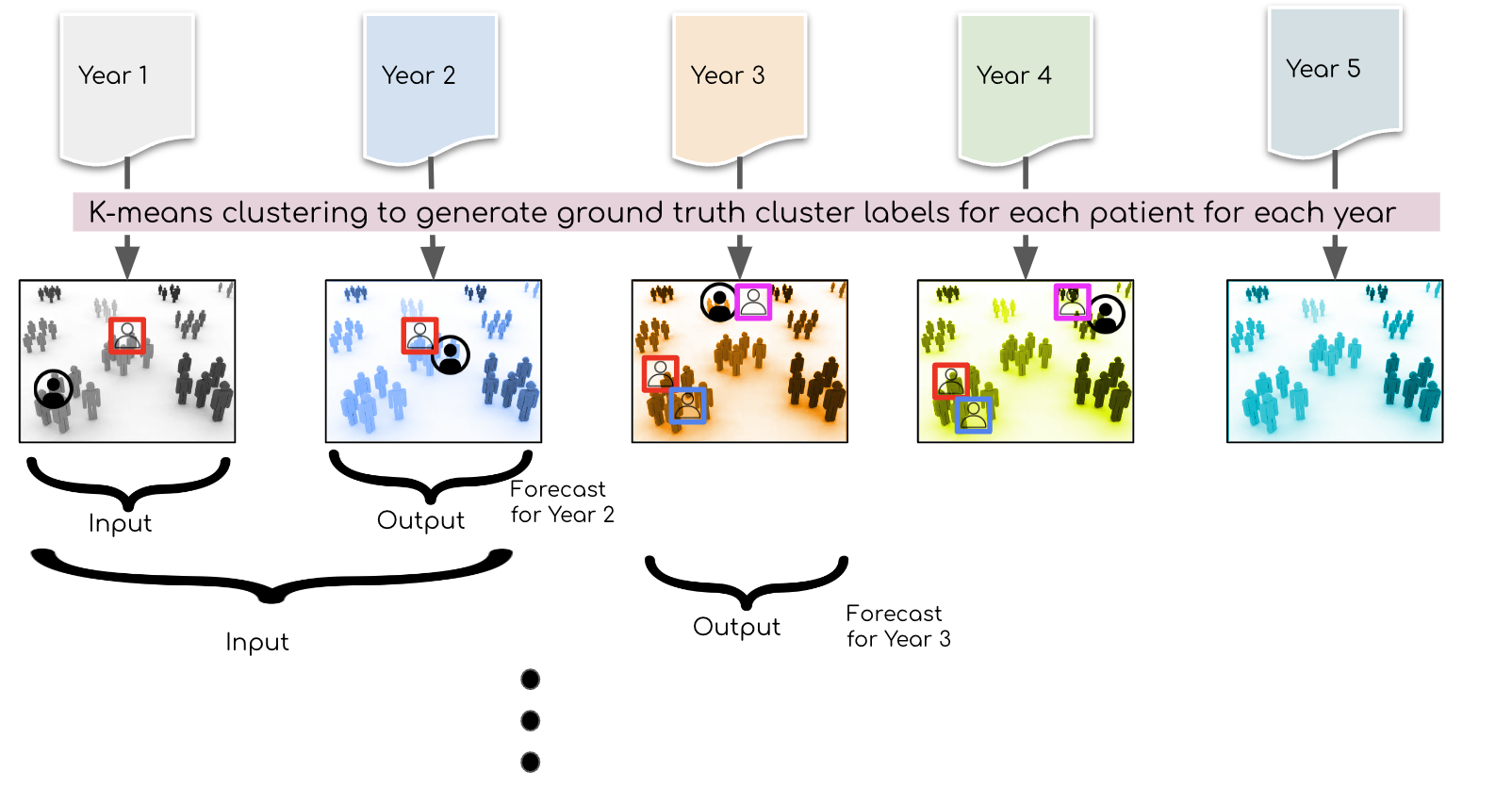}
    \caption{An illustration of our predictive clustering for yearly patient phenotype forecasting (Input is the vital signs data and cluster labels for a year, Output is the cluster assignment forecast for a future year).}
    \label{fig:forecast}
\end{figure*}

In this paper, we address the problem of estimating future subjective pain scores in patients with SCD using six objective physiological measures collected by nurse coordinators in a hospital. We propose a predictive clustering based approach \cite{lee2020temporal} to combine predictions on the future outcomes with clustering as illustrated in Figure \ref{fig:forecast}. 

\textbf{Illustration of Proposed Approach for Long-term Pain Management:} We divide the entire data into  bins corresponding to the year. In this case, we divide the dataset into five bins as we had data from five consecutive years. Next, we apply clustering algorithm on each of the bins to generate ground truth cluster labels for each patient in each year. As shown in Figure \ref{fig:forecast}, two patients- black circle and red square belong to two different clusters in year 1. However, they might behave in a similar pattern in year 2 and get clustered together. Again, in year 3, their disease progression (vitals, pain) might be different and they are clustered separately. For long-term chronic pain management, we propose to forecast these future cluster alignments of patients. A clinician can utilize the information about such temporal patient phenotyping to design appropriate pain medication prescription for a patient so as to treat the patient and avoid any risks of addiction to pain medications.

We utilize 51718 time-series Electronic Health Records (EHR) data points from 498 participants at Duke University Hospital over five consecutive years. As the dataset had one or more missing values for each time stamp, we first design a clinically motivated data interpolation method. Then we evaluate supervised and self-supervised machine learning algorithms for short-term pain forecasting and proposed long-term patient phenotype forecasting. Our results show that a variational autoencoder based learning method outperforms other methods in both the cases. In summary, we make the following contributions:
\begin{itemize}
    \item We propose a clinically motivated data interpolation approach considering the irregularity in clinical visits and inconsistencies in data records.
    \item We show that self-supervised learning-based methods perform best while forecasting future pain at short-term (hourly) duration.
    \item We propose a patient phenotyping approach based on dynamic time warping distance and self-supervised learning for long-term (yearly) patient subgroup/profile forecasting.
    \item We show in a case study that our self-supervised patient phenotyping approach is able to capture the interplay between multiple vital signs influencing the evolution of future patient pain profiles.
\end{itemize}
While the proposed approach is applied to patients with SCD, it can be readily applied to other chronic diseases such as diabetes, or cystic fibrosis due to its similar data structure.

\section{Related Work}
Recently, studies on pain forecasting are gaining attention. There have been attempts to forecast pain, specifically postoperative pain, using data other than physiology and activity measurements. Tighe et al. \cite{tighe2015teaching} explored various classification algorithms to forecast whether a patient was at risk for moderate to severe postoperative pain for postoperative day 1 and day 3 using 796 clinical variables from Electronic Medical Records (EMRs) in a retrospective cohort of 8,071 surgical patients. In forecasting moderate to severe postoperative pain for the postoperative day (POD) 1, the LASSO algorithm, using all 796 variables, had the highest accuracy with an area under the receiver-operating curve (ROC) of 0.704. Next, the gradient-boosted decision tree had a ROC of 0.665, and the k-nearest neighbors (k-NN) algorithm had a ROC of 0.643. For POD 3, the LASSO algorithm, using all variables, again had the highest accuracy, with a ROC of 0.727. Logistic regression had a lower ROC of 0.5 for predicting pain outcomes on POD 1 and 3. The same group also developed a model based on RNNs to forecast pain levels after administering specific pain medication and trained it on pain score patterns \cite{nickerson2016deep}. 

Pain forecasting is a critical issue in pain management in SCD. Based on the recent advancements in the field and previous works from our group \cite{yang2020improving}, we believe designing data-driven self-supervised learning models based on physiological data is a promising approach for pain forecasting. 
Temporal clustering, also known as time-series clustering,
is a process of unsupervised partitioning of the time-series
data into clusters in such a way that homogeneous time-series are grouped together based on a certain similarity
measure. Temporal clustering is challenging because i) the
data is often high-dimensional it consists of sequences not
only with high-dimensional features but also with many
time points and ii) defining a proper similarity measure
for time-series is not straightforward since it is often highly
sensitive to distortions \cite{ratanamahatana2005novel}. To
address these challenges, there have been various attempts
to find a good representation with reduced dimensionality
or to define a proper similarity measure for times-series
\cite{aghabozorgi2015time}. 

Recently, \cite{baytas2017patient} and \cite{madiraju2018deep}
proposed temporal clustering methods that utilize low dimensional representations learned by recurrent neural networks (RNNs). These works
are motivated by the success of applying deep neural networks to find ``clustering friendly" latent representations for
clustering static data \cite{xie2016unsupervised, yang2017towards}. In
particular, Baytas et al. \cite{baytas2017patient} utilized a modified LSTM
auto-encoder to find the latent representations that are effective to summarize the input time-series and conducted
K-means on top of the learned representations as an adhoc process. Similarly, Madiraju et al. \cite{madiraju2018deep} proposed
a bidirectional-LSTM auto-encoder that jointly optimizes
the reconstruction loss for dimensionality reduction and the
clustering objective. However, these methods do not associate a target property with clusters and, thus, provide little
prognostic value about the underlying disease progression. Recently, Lee et.al. \cite{lee2020temporal} have addressed this issue by proposing a temporal predictive clustering approach. Along similar direction, we design our long-term patient phenotype forecasting.

\section{Data Preparation}
For chronic diseases, it is expected that the knowledge discovery in clinical time-series data will play a potent role. The reason is that it is humanly impossible even for experienced medical doctors to directly extract the tendencies of a great deal of multivariate time series. In order to obtain knowledge on changes in symptom, time-series analysis should be applied to clinical time-series data. However, the unfavorable characteristics of clinical time series data makes it impossible to apply conventional time series analysis methods that assume regular sampling. Patient’s clinical examination records basically have irregular intervals, which are caused by irregular visits of the patient for clinical examination and the intensive execution of clinical examination when the patient’s illness becomes worse. It is needed to make the intervals of clinical time-series data equal by interpolating them to for the application of the time series analysis methods. The simplest interpolation methods are averaging or linear regression with a constant period \cite{ohsaki2002rule}. These methods slide a time window of constant width and calculate the average or the estimate by linear regression on the points in each window. They regard it as an interpolated point, namely a set of an equally spaced time point and the estimated value corresponding to the time point. Consequently, the time intervals are made equal via this process. However, the symptoms are different among patients and non-stationary for each patient, these methods thus have some problems, how the width and overlap of a time window are determined and whether it is proper to fix the width and overlap. We therefore propose a clinically-motivated interpolation method for this purpose in a step method: (1) First, we extracted the visit information of each record in our dataset following the definitions by Padhee et. al. \cite{padhee2020pain}, (2) Second, we modified the time grid rounding up to the nearest hour, and (3) Third, within each visit, we applied linear interpolation within 2 hours window based on the clinical relevance advice from our co-author clinician. 

\begin{table}[]
\centering
\caption{Percentage of missing data before interpolation (raw data) and after interpolation.}
\label{forecastmissing}
\begin{tabular}{|l|l|l|}
\hline
\textbf{Variable} & \textbf{Raw Data} & \textbf{After Interpolation} \\ \hline
BP                & 74.481765         & 37.137940                    \\ \hline
SpO2              & 65.178842         & 3.758846                     \\ \hline
Pulse             & 67.738294         & 0.425384                     \\ \hline
Resp              & 67.274355         & 1.674465                     \\ \hline
Temp              & 78.210408         & 2.225531                     \\ \hline
Pain Score        & 68.536177         & 12.732511                    \\ \hline
\end{tabular}
\end{table}

In this study, we utilized 51718 records from 498 participants at Duke University Hospital over a maximum of five consecutive years. Each record contained measures for six vital signs as follows: (i) peripheral capillary oxygen saturation (SpO2), (ii) systolic blood pressure (SystolicBP), (iii) diastolic blood pressure (DiastolicBP), (iv) heart rate (Pulse), (v) respiratory rate (Resp), and (vi) temperature (Temp). Along with the vital signs, each record also included the patient's self-reported pain score with an ordinal range from 0 (no pain) to 10 (severe and unbearable pain). The data were de-identified using study labels to label the patient without identification. The timestamp for each data entry was also de-identified, preserving temporality. The dataset had missing values for one or more of the vital signs and the pain score. Table \ref{forecastmissing} shows that we reduced the statistics of missing variables in our dataset significantly, while preserving the clinical semantics.

\section{EXPERIMENTS AND RESULTS}
We design our experiments in a two-stage framework consisting of 1) short-term pain forecasting, and 2) long-term patient phenotype forecasting. In this section, we discuss both the modules in detail.

\subsection{Short-term pain forecasting}
We design two scenarios to evaluate multiple supervised and self-supervised machine learning algorithms for short-term pain forecasting: (1) Individualized patient scenario, and (2) Mixed patient scenario. In an individualized patient scenario, we included past and future sequences from the same patient as input and output. In a mixed patient scenario, we combined all the patient data. A past sequence from a patient is input data and
output is the forecast on a future sequence from a random patient, including the patient in the input (training) set. We implemented six models to predict future pain score: (1) Baseline Random Forest (RF) Regression model, (2) Time-series Autoregressive Integrated Moving Average model, (3) Dense Regression model (Multilayer Perceptron model), (4) Long Short-Term Memory model, (5) Contrastive Predictive Coding (CPC) with RF regression model, and (6) Variational Autoencoder (VAE) with RF regression model. 

\begin{figure*}
    \includegraphics[width=\textwidth,scale=0.7]{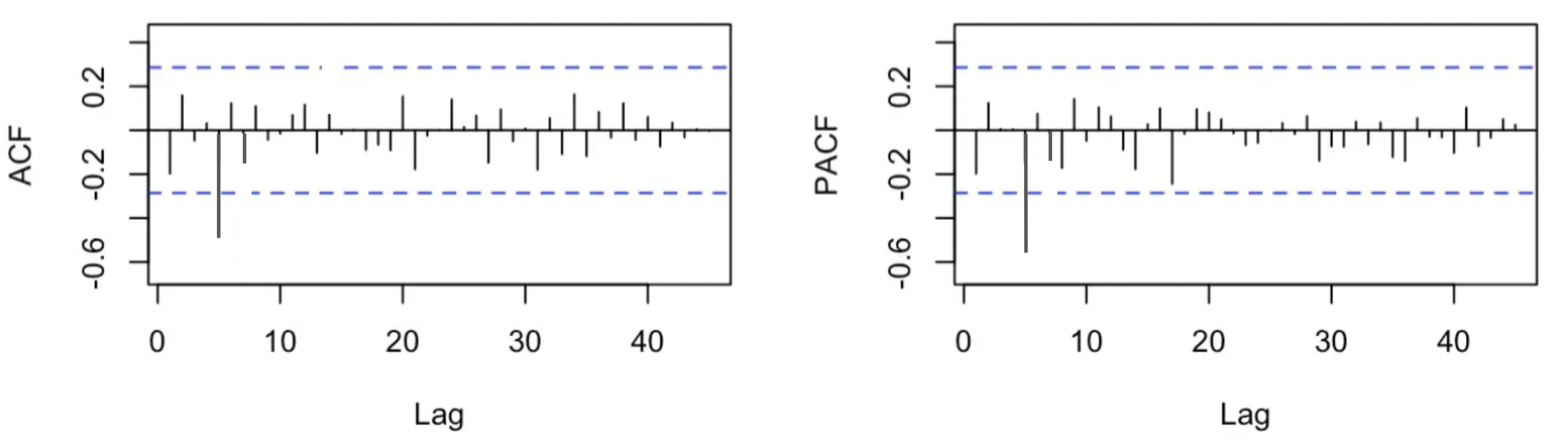}
    \caption{Initial Autocorrelation Function (ACF) and Partial Autocorrelation Function (PACF) plots showing significant lag at 5.}
    \label{acf}
\end{figure*}

\begin{figure*}

    \includegraphics[width=\textwidth,scale=0.7]{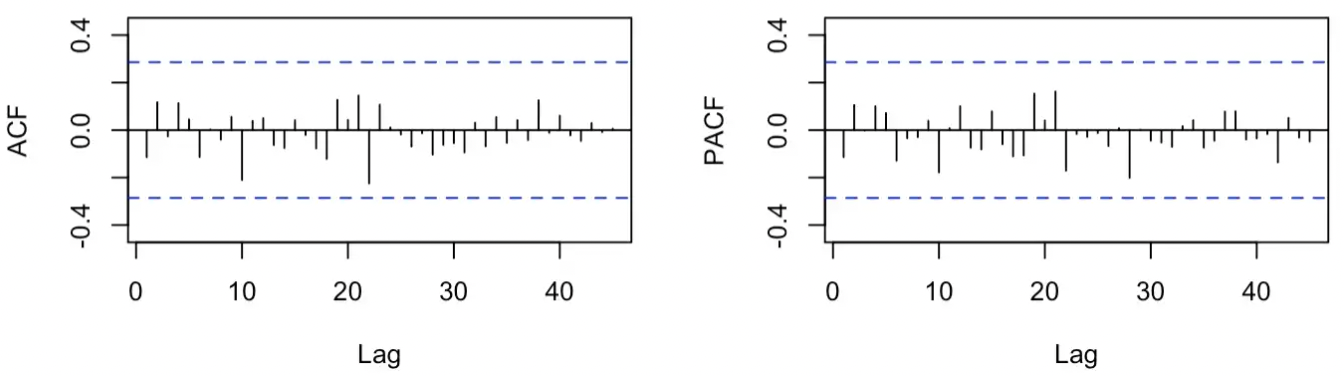}
    \caption{Final model Autocorrelation Function (ACF) and Partial Autocorrelation Function (PACF) plots showing no significant lag.}
    \label{acf2}
\end{figure*}

An ARIMA model has three parameters p, q, and d. Significant lags  at 5 in the autocorrelation and partial autocorrelation function plots (Figure \ref{acf}) extend beyond the dashed blue lines and indicate poor model fit. We decided the p and q parameters range for the grid search as 0 and 5. Thus, we applied a grid to search by passing the integer values in the range [0, 5] for both p and q and decided on the value of the p and q parameters to be 1 with AIC value of the model as 584.9 and both ACF and PACF plots showed no significant lags (Figure \ref{acf2}). Based on the ADF test \cite{dickey1981likelihood}, we found the time series of pain score to be non-stationary and required one-time differencing (d = 1). 

MLP and LSTM models were trained on physiology and pain scores on the interpolated time series dataset. We adjusted the hyper-parameters (layers, neurons, batch size, and epochs) in MLP and LSTM models to have an idea of the range (minimum and maximum) value of the parameters. The hyperparameters higher and lower than the values for which we did not get any improvements in the model’s fit provided us with the minimum and the maximum range. Next, we varied the hyper-parameters in the obtained range and evaluated the model performance. We validated the model training on test dataset. Since we did not want to provide the memory-based
LSTM model an advantage over the memory-less MLP model; we tested the same range of hyper-parameters for both the MLP and LSTM models. After getting the best set of hyper-parameters, we evaluated the model performance on these best hyper-parameters 40 times as there is a run-to-run variability in the model’s output on training data. We finalized the model with the least objective function value among the 40 model runs as the final prediction from the model. For training both models (MLP and LSTM), we used the ReLU activation function. These models were created in Keras using the Tensorflow backend. 

In self-supervised learning, the network is trained to predict future physiological data from extensive unlabeled past physiological data. During the training process, our self-supervised learning network learned latent representations that were used to infer future pain states using a regression model. We used a similar architecture by Yang et.al. \cite{yang2020improving} to learn representations from physiological signals using a CPC network and a VAE architecture by our previous work \cite{padhee2022improving}. Specifically, we used a three-layer Convolutional Neural Network (CNN) \cite{lecun2015deep} as the encoder in CPC model. We then used a gated recurrent unit (GRU) based Recurrent Neural Network (RNN) \cite{lecun2015deep} for the autoregressive part of the model with 64 dimensional of hidden states. The output of the GRU based RNN model $c_t$ is then used as the feature for pain forecasting task. The pretext task network was trained using the Adam optimizer with a batch size of 128 and a learning rate of $10^{-3}$. The network structure and hyperparameters were tuned based on experiments to maximize the accuracy of the pretext task.

Next, we trained a regression model to predict future pain values using the representations learned (the output of the CPC and VAE network) as input features. To summarize, we used a past physiological signal sequence in the trained self-supervised network to generate the latent representation, which is used as the input feature of a regression model to predict the pain score reported at a future time step. In the downstream task, we trained a regression model to predict future pain values using the latent context representation $c_t$ (the output of the autoregressive network) as input features. Specifically, a past vital signs sequence was fed into the trained CPC network to generate the context latent representation $c_t$. Then $c_t$ was used as the input feature of a regression model to predict raw (not interpolated) pain scores reported at a future time step. Due to the lack of pain score labels (as we used raw pain scores instead of the interpolated pain scores for prediction), we utilized random forest \cite{breiman2001random} as the supervised regression model for pain forecasting. We chose this model because ensemble methods are more robust and have advantages in dealing with small sample sizes \cite{yang2010review}. 

\subsection{Long-term patient phenotype forecasting}
Clustering is an unsupervised learning process where an algorithm brings similar data points closer without any ``ground truth” labels. The similarity between data points is measured with a distance metric, commonly Euclidean distance. In general, the Euclidean distance metric (or other types of Minkowski metric) is used to find an average of all the data within the clusters. However, its one-to-one mapping nature cannot capture the average shape of the two time series, in which case Dynamic Time Warping (DTW) \cite{zhang2022dynamic} is more favorable. 

Clustering different time series data is challenging as each data point acts as an ordered sequence. Classically, the most common approach involves flattening the time series (sequence) into a table, with a column for each time point (or an aggregation of the entire sequence), and applying standard clustering algorithms like k-means. 
However, these clustering algorithms use standard measures such as Euclidean distance, which is often not the best for time series (ordered sequences). Hence, we replace the default distance measure with DTW to compare the temporal sequences that can measure the similarity between two sequences that do not align with each other rigidly in time, speed, or length. 
Unlike the Minkowski distance function, dynamic time warping breaks the one-to-one alignment limitation and supports non-equal-length time series. It uses a dynamic programming technique to find all possible paths and selects the one that yields a minimum distance between the two sequences of time series using a distance matrix, where each element in the matrix is a cumulative distance of the minimum of the three surrounding neighbors. Suppose we have two time series, a sequence $Q = q_1, q_2, …, q_i, …, q_n$ and a sequence $C = c_1, c_2, …, c_j, …, c_m$. First, we create an $n$ x $m$ matrix, where every $(i, j)$ element of the matrix is the cumulative distance of the distance at $(i, j)$ and the minimum of the three elements neighboring the $(i, j)$ element, where $0< i\leq n$ and $0< j\leq m$. We can define the $(i, j)$ element as:

\begin{equation} \label{dtw1}
e_{ij} = d_{ij} + min\left \{ e_{(i-1)(j-1)}, e_{i(j-1)}, e_{(i-1)j} \right \}
\end{equation}

where $d_ij = (c_i + q_j)^2$ and $e_ij$ is $(i, j)$ element of the matrix which is the summation between the squared distance of $q_i$ and $c_j$, and the minimum cumulative distance of the three elements surrounding the $(i, j)$ element. Then, to find an optimal path, we have to choose the path that gives minimum cumulative distance at $(n, m)$. The distance is defined as:

\begin{equation} 
D_{DTW}\left ( Q,C \right ) = \underset{\forall w\in P}{min}\left \{ \sqrt{\sum_{k=1}^{K}} d_{wk}\right \}
\label{dtw2}
\end{equation}
where $P$ is a set of all possible warping paths, and $wk$ is $(i, j)$ at $k^th$ element of a warping path and $K$ is the length of the warping path.

We compute the cluster centroids with respect to DTW by minimizing the sum of squared DTW distance between the cluster centroid and the series in the cluster. We employ k-means clustering for each year of patient data and generated cluster labels for all patients for each year. Next, we use the cluster labels as an additional feature to our pain forecasting models to predict the future cluster label (ground truth) for next year.

\section{Results}
Tables \ref{res:forcast1} and \ref{res:forcast2} show the MAE and $R^2$ for forecasting pain scores for mixed patients and individual patient models respectively. We present the results using the best predictions from 40 runs from each model. For mixed patient forecasting, we combined all the patient data. The training data consisted of a past sequence from a patient, and we generated the forecast on a future sequence from a random patient, including the patient in the training set. In the individualized patient experiment, we included past and future sequences from the same patient. 

Next, for each year, we used K means clustering using DTW distance on all patients' physiology and pain scores data to obtain an optimal number of seven clusters [with a Normalized Mutual Information (NMI) score of 0.35, purity score of 0.67, Silhouette Index of 0.12]. We treat the cluster labels generated as ground truth labels for each patient for each year. Our goal is to understand if we can predict future cluster alignment of patients. Hence, we train the same models discussed above on a past sequence of physiology data, pain scores, and cluster labels to forecast the cluster label for the following year. We report the AUROC of our models with raw pain scores as test data for years 2,3,4 and 5 in Table \ref{res:forcast3}. The training for each is the data for all previous years, i.e., for the forecast of year 5, we trained models on interpolated data from years 1,2,3, and 4 and tested on original data available for year 5. 

\begin{table}[]
\centering
\caption{Pain forecasting results for individualised patient model}
\label{res:forcast1}
\scalebox{0.8}{
\begin{tabular}{|l|ll|ll|ll|}
\hline
\multirow{2}{*}{}                  & \multicolumn{2}{l|}{1 hour}                               & \multicolumn{2}{l|}{2 hour}                                & \multicolumn{2}{l|}{4 hour}                                \\ \cline{2-7} 
                                   & \multicolumn{1}{l|}{MAE}           & $R^2$ & \multicolumn{1}{l|}{MAE}            & $R^2$ & \multicolumn{1}{l|}{MAE}            & $R^2$ \\ \hline
\textbf{VAE + RF Regression Model} & \multicolumn{1}{l|}{\textbf{1.16}} & \textbf{0.91}        & \multicolumn{1}{l|}{\textbf{1.295}} & \textbf{0.91}        & \multicolumn{1}{l|}{\textbf{1.484}} & \textbf{0.90}        \\ \hline
CPC + RF Regression Model          & \multicolumn{1}{l|}{1.184}         & 0.89                 & \multicolumn{1}{l|}{1.369}          & 0.89                 & \multicolumn{1}{l|}{1.57}           & 0.88                 \\ \hline
LSTM                               & \multicolumn{1}{l|}{1.261}         & 0.84                 & \multicolumn{1}{l|}{1.482}          & 0.84                 & \multicolumn{1}{l|}{1.663}          & 0.84                 \\ \hline
Dense Regression (MLP)             & \multicolumn{1}{l|}{1.294}         & 0.79                 & \multicolumn{1}{l|}{1.342}          & 0.79                 & \multicolumn{1}{l|}{1.47}           & 0.78                 \\ \hline
ARIMA                              & \multicolumn{1}{l|}{1.63}          & 0.74                 & \multicolumn{1}{l|}{1.64}           & 0.73                 & \multicolumn{1}{l|}{1.70}           & 0.73                 \\ \hline
RF Regression Model                & \multicolumn{1}{l|}{1.85}          & 0.62                 & \multicolumn{1}{l|}{1.86}           & 0.63                 & \multicolumn{1}{l|}{1.90}           & 0.61                 \\ \hline
\end{tabular}
}
\end{table}

The best MAE (= 0.58) on test data was obtained for our LSTM based VAE model which contained 2 hidden layers, 4 neurons in each hidden layer, 20 batch size, and for 30 epochs. In general, all the models resulted in lower MAE and higher $R^2$ scores in individualised models. As seen in the tables, both the MLP and LSTM models outperformed the RF Regression model (baseline) as well as the ARIMA model. Also, the self-supervised LSTM based VAE model performed the best among all the models. 

\begin{table*}

\centering
\caption{Pain forecasting results for mixed patient model}
\label{res:forcast2}
\scalebox{0.8}{
\begin{tabular}{|l|ll|ll|ll|}
\hline
\multirow{2}{*}{}                                           & \multicolumn{2}{l|}{1 hour}                                          & \multicolumn{2}{l|}{2 hour}                                          & \multicolumn{2}{l|}{4 hour}                                         \\ \cline{2-7} 
                                                            & \multicolumn{1}{l|}{MAE}                      & $R^2$ & \multicolumn{1}{l|}{MAE}                      & $R^2$ & \multicolumn{1}{l|}{MAE}                     & $R^2$ \\ \hline
\textbf{VAE + RF Regression Model \cite{padhee2022improving}} & \multicolumn{1}{l|}{\textbf{0.58 (+/- 0.39)}} & \textbf{0.91}        & \multicolumn{1}{l|}{\textbf{0.63 (+/- 0.42)}} & \textbf{0.89}        & \multicolumn{1}{l|}{\textbf{0.78(+/- 0.42)}} & \textbf{0.82}        \\ \hline
CPC + RF Regression Model  \cite{yang2020improving}         & \multicolumn{1}{l|}{0.76 (+/-0.54)}           & 0.90                 & \multicolumn{1}{l|}{0.79 (+/- 0.54)}          & 0.88                 & \multicolumn{1}{l|}{0.96 (+/- 0.58)}         & 0.88                 \\ \hline
LSTM                                                        & \multicolumn{1}{l|}{0.78 (+/-0.66)}           & 0.88                 & \multicolumn{1}{l|}{0.73 (+/- 0.62)}          & 0.87                 & \multicolumn{1}{l|}{0.98 (+/- 0.65)}         & 0.87                 \\ \hline
Dense Regression (MLP)                                      & \multicolumn{1}{l|}{1.01 (+/- 0.75)}          & 0.76                 & \multicolumn{1}{l|}{1.142 (+/- 0.78)}         & 0.76                 & \multicolumn{1}{l|}{1.27 (+/- 0.84)}         & 0.76                 \\ \hline
ARIMA                                                       & \multicolumn{1}{l|}{1.24 (+/-0.87)}           & 0.65                 & \multicolumn{1}{l|}{1.28 (+/-0.91)}           & 0.61                 & \multicolumn{1}{l|}{1.45 (+/-0.93)}          & 0.63                 \\ \hline
RF Regression Model                                         & \multicolumn{1}{l|}{1.37 (+/-0.89)}           & 0.62                 & \multicolumn{1}{l|}{1.42 (+/-0.91)}           & 0.61                 & \multicolumn{1}{l|}{1.53 (+/-0.94)}          & 0.62                 \\ \hline
\end{tabular}
}
\end{table*}

\begin{table}[]
\centering
\caption{Predictive Clustering for Long-term pain forecasting (AUROC)}
\label{res:forcast3}
\begin{tabular}{|l|l|l|l|l|}
\hline
         & Year 2 & Year 3 & Year 4 & Year 5 \\ \hline
MLP      & 0.633  & 0.691  & 0.729  & 0.775  \\ \hline
CPC + RF & 0.721  & 0.793  & 0.852  & 0.886  \\ \hline
\textbf{VAE + RF} & \textbf{0.743} & \textbf{0.832} & \textbf{0.893} & \textbf{0.921} \\ \hline
\end{tabular}
\end{table}

We report the area under the receiver operating characteristic (AUROC) score for evaluating our long-term cluster forecasting models. Also, we compare the performance of our best performing models from short-term forecasting as shown in Table \ref{res:forcast3}. First, as expected, our LSTM based self supervised VAE netowrk performed best in long-term pain forecasting. Second, each model performed best in forecasting the cluster assignment for year 5 as it had more data to learn from (year 1-4). 

\section{Discussion OF RESULTS}
The results presented in the previous section led to the below findings. The primary objective of this research was short-term pain forecasting and evaluating the performance of existing statistical (ARIMA), supervised neural (MLP and LSTM), and self-supervised (CPC, VAE) models for individual and mixed patient scenarios. Another objective of this paper was to systematically evaluate a predictive clustering based approach for long-term pain forecasting. Overall, we expected the self-supervised models to perform better than the statistical and supervised neural models. First, as per our expectation, the best performance in terms of error was found from the VAE trained network, followed by CPC trained network, LSTM, MLP, ARIMA and RF regression models. A likely reason for these results is that ARIMA models are perhaps not able to capture the non-linearities present in the time-series data. Thus, overall, these models tended to perform not as well compared to other models. Also, overall, the neural network models (MLP and LSTM) performed similarly and better than the persistence and ARIMA models. That is likely because our dataset is were non-linear and neural network models, by
their design, could account for the non-linear trends in datasets. However, another reason for this result could be simply because the self-supervised network models possess several weights (parameters), whereas the ARIMA model possesses only three parameters. 

\subsection{Changing Patient Phenotype over time : Long-term forecasting}
\begin{figure}
    \includegraphics[width=3.5in]{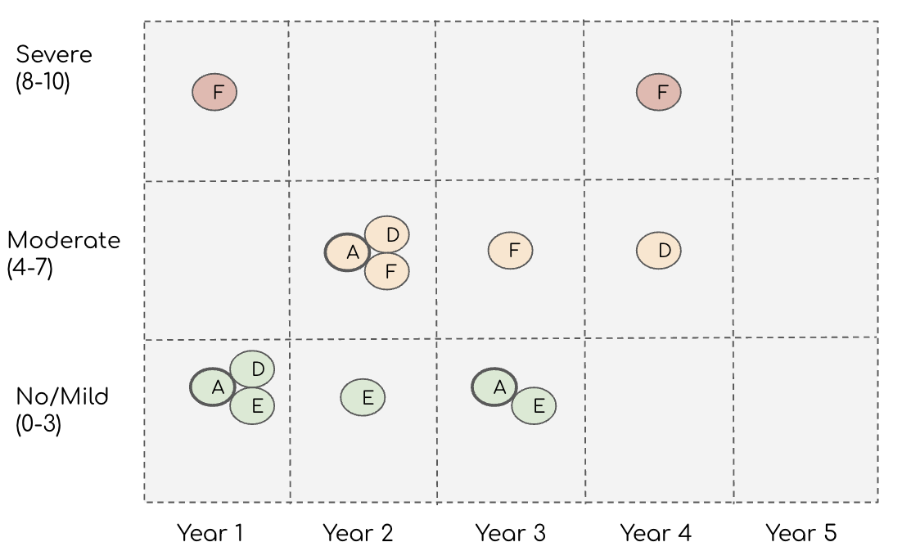} 
    \caption{An illustration of run-time examples of our clustering based long term pain forecasting on six representative patients.}
    \label{cp}
    
\end{figure}

In this subsection, we demonstrate run-time examples of how our predictive clustering approach was able to flexibly update the cluster assignments over time with respect to the future pain in the next year. We present a case study of six representative patients as discussed below and shown in Figure \ref{cp}. 

\begin{figure}
    \includegraphics[width=3.5in]{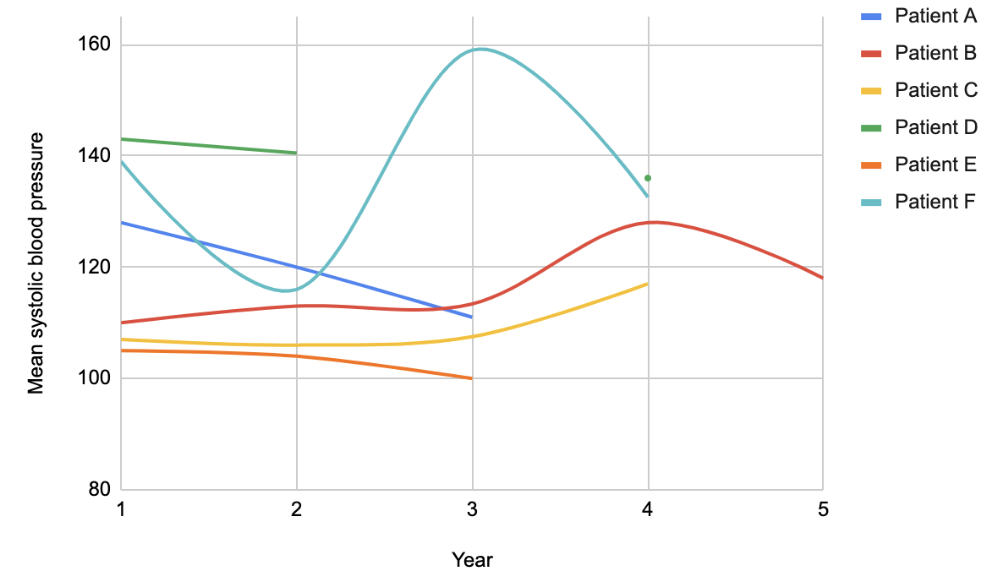} 
    \caption{Change in systolic blood pressure (mean) across years for the six representative patients.}
    \label{bp}
    
\end{figure}

In the second year, he/she had moderate average pain (pain score 4-7). Our clustering model was able to predict the temporal phenotype assigned to this patient as similar to that of patient F who had moderate average pain (pain score 4-7). As shown in Figure \ref{bp}, we see that the systolic blood pressure for both these patients follow a similar trend (decreasing). Furthermore, our clustering algorithm phenotyped patient A and patient E together in the third year to a cluster predominantly having low/mild pain scores. Both of them had mild pain in the first year. So, our approach could change the phenotype of patient A from low/mid pain to moderate pain and again back to low/mild pain. We can see from Figure \ref{bp} that patients A and E followed a decreasing trend in systolic blood pressure from first year till third year. Furthermore our model predicted accurately patient D to be in the same cluster with patient A and F having moderate pain and decreasing blood pressure in the second year. Patient D had mild pain in first year, and moderate pain in the second year. 

\textbullet\ Patient B had an average moderate pain score in the first year and maintained a moderate pain score in second year, no pain in third and fourth year, and an average moderate pain in the fifth year. Our clustering model predicted that patient B and patient C belonged to the same cluster in second year (both had moderate pain). In the third year also, they were clustered together although patient B had no pain and patient C had moderate pain. We analyzed from the Figure \ref{bp} that although belonging to different pain range, they followed a similar trend in systolic blood pressure. Interestingly, our algorithm then allocated them to separate clusters in the fourth year. While patient B was allocated to a cluster with mixed pain scores, patient C was clustered with patients with moderate pain score.

\subsection{Limitations}

Based on our observations, we would like to point out a limitation in predictive clustering, the trade-off between the clustering performance (for better interpretability) – which quantifies how the data samples are homogeneous within each cluster and heterogeneous across clusters with respect to the future outcomes of interest – and the prediction performance is a common issue. The most critical parameter that governs this trade-off is the number of clusters. More specifically, increasing the number of clusters will give the predictive clusters higher diversity to represent the output distribution and, thus, increase the prediction performance while decreasing the clustering performance. One extreme example is that there are as many clusters as data samples which will make the identified clusters fully individualized; consequently, each cluster will lose interpretability as it no longer groups similar data samples.

\section{CONCLUSION}
This paper proposes a self-supervised learning method for short-term and long-term pain forecasting. We evaluated the performance of supervised, self-supervised, and statistical time-series techniques for performing short-term (hourly) and long-term (yearly) time-series predictions of longitudinal healthcare data. Throughout the experiments, we showed that our self-supervised model achieves superior performance over state-of-the-art methods and identifies meaningful clusters that can be translated
into actionable information for clinical decision-making. This work can be extended by incorporating other biological, social, and psychological factors.

\section*{Acknowledgment}
The authors would like to thank the the National Institutes of Health for support through grant R01AT010413.

\bibliographystyle{IEEEtran}
\bibliography{IEEEabrv,sample}
\end{document}